\documentclass[10pt,twocolumn,letterpaper]{article}

\usepackage{cvpr}              

\usepackage[dvipsnames]{xcolor}

\newcommand{\todo}[1]{{\color{red}#1}}

\usepackage{caption}
\usepackage{subcaption}
\usepackage{graphicx}
\usepackage{amsmath}
\usepackage{amssymb}
\usepackage{booktabs}
\usepackage{enumitem}
\usepackage{color}
\usepackage{multicol}
\definecolor{cvprblue}{rgb}{0.21,0.49,0.74}

\usepackage[pagebackref,breaklinks,colorlinks,linkcolor=cvprblue,urlcolor=cvprblue,citecolor=cvprblue]{hyperref}
\definecolor{citecolor}{HTML}{0071BC}
\definecolor{linkcolor}{HTML}{ED1C24}

\usepackage[capitalize]{cleveref}
\crefname{section}{Sec.}{Secs.}
\Crefname{section}{Section}{Sections}
\crefname{table}{Tab.}{Tabs.}
\Crefname{table}{Table}{Tables}

\definecolor{poscolor}{RGB}{212,17,89}
\definecolor{color2}{RGB}{250,130,130}
\definecolor{color3}{RGB}{130,130,130}
\definecolor{color4}{RGB}{14,83,246}

\newcommand{\ourmethod}{UniGS}

\newif\ifdrafting
\draftingtrue 
\ifdrafting
    \newcommand{\diff}[2]{{\color{red}\sout{#1}} {\color{green}#2}}
    \newcommand{\missing}[1]{{\color{orange}{#1}}}
\else
    \newcommand{\todo}[1]{}
    \newcommand{\diff}[1]{}
    \newcommand{\missing}[1]{}
\fi

\usepackage{algorithm}
\usepackage{bm}
\usepackage{algorithmic}
\usepackage{listings}
\usepackage{makecell}  
\usepackage{diagbox} 
\usepackage{mathtools} 
\usepackage{graphicx} 
\usepackage{multirow}
\usepackage{nth}
\usepackage{amssymb}
\usepackage{pifont}
\usepackage{listings}
\usepackage{xcolor}

\newcommand{\cmark}{\ding{51}}%
\newcommand{\xmark}{\ding{55}}%

\def\paperID{} 
\def\confName{}
\def\confYear{}

\title{UniGS: Unified Representation for Image Generation and Segmentation}

\author{
Lu Qi$^{1}$\thanks{Equal contribution. $\dagger$ corresponding author.},
~~~
Lehan Yang$^{2*}$,~~~
Weidong Guo$^{3\dagger}$,~~~
Yu Xu$^3$,~~~ \\
Bo Du$^{4}$,~~~
Varun Jampani$^5$, ~~~
Ming-Hsuan Yang$^{1,6}$, ~~~
\\[0.2cm]
$^1$The University of California, Merced~~
$^2$The University of Sydney~~ \\
$^3$QQ Brower Lab, Tencent,~~
$^4$Wuhan Univeristy~~ \\
$^5$Stability AI~~
$^6$Google Research~~
}

\begin{document}
\maketitle
\begin{abstract}
This paper introduces a novel unified representation of diffusion models for image generation and segmentation. Specifically, we use a colormap to represent entity-level masks, addressing the challenge of varying entity numbers while aligning the representation closely with the image RGB domain. Two novel modules, including the location-aware color palette and progressive dichotomy module, are proposed to support our mask representation. On the one hand, a location-aware palette guarantees the colors' consistency to entities' locations. On the other hand, the progressive dichotomy module can efficiently decode the synthesized colormap to high-quality entity-level masks in a depth-first binary search without knowing the cluster numbers. To tackle the issue of lacking large-scale segmentation training data, we employ an inpainting pipeline and then improve the flexibility of diffusion models across various tasks, including inpainting, image synthesis, referring segmentation, and entity segmentation. Comprehensive experiments validate the efficiency of our approach, demonstrating comparable segmentation mask quality to state-of-the-art and adaptability to multiple tasks. The code will be released at \href{https://github.com/qqlu/Entity}{https://github.com/qqlu/Entity}.
\end{abstract}    
\section{Introduction}
\label{sec:intro}
Deep learning has propelled the performance of several
tasks to new heights, marking substantial progress within the computer vision community. Image generation~\cite{gregor2015draw,li2019controllable,zhao2019image,li2022mat} and segmentation~\cite{chen2017deeplab,zhao2017pyramid,he2017mask,liu2018path,qi2022open,qi2023high}, as two typical dense prediction tasks within this field, are widely used in plethora of applications such as autonomous driving~\cite{qi2019amodal}, video surveillance~\cite{shu2014human}, medical imaging~\cite{suetens2017fundamentals}, robotics~\cite{chu2021icm}, photography~\cite{tsai2023dual}, and intelligent creation~\cite{wang2021image,wang2022palgan}.

\begin{figure}[!tbp]
\centering
\includegraphics[width=0.47\textwidth]{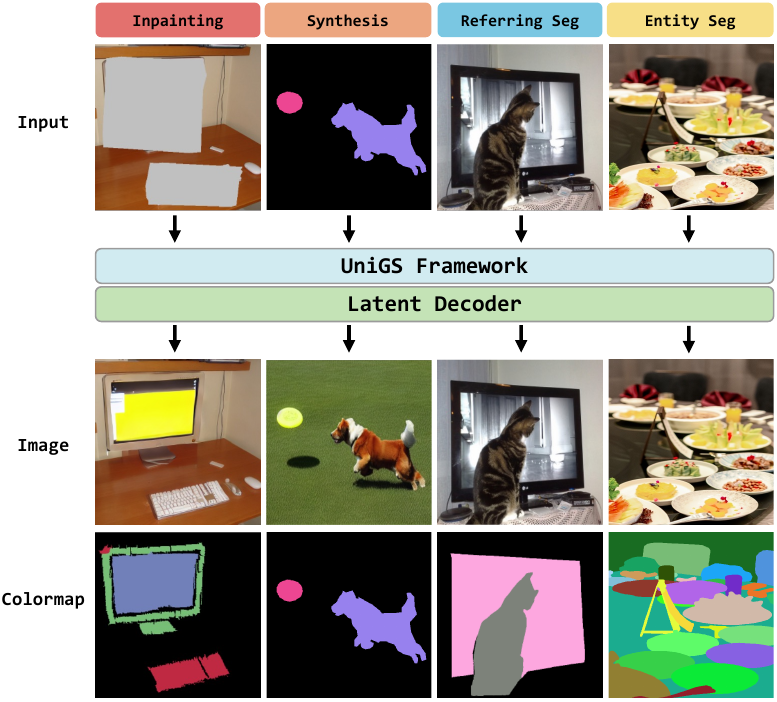}
\caption{\textbf{Visualization results of a single UniGS model on image generation and segmentation.} We present four tasks: multi-class multi-region inpainting, image synthesis, referring segmentation, and entity segmentation. We note that the generation of colormaps shares a similar pipeline with images without needing any explicit segmentation loss.}
\label{fig.teaser}
\vspace{-2mm}
\end{figure}

The innovative usage of latent codes~\cite{rombach2022high} in diffusion models has recently demonstrated remarkable capabilities in producing high-quality images, opening a new era of AI-generated content (AIGC). Nevertheless,
using a similar design for segmentation remains relatively unexplored in diffusion-based works, despite evidence from specific studies~\cite{chefer2023attend,burgert2022peekaboo,liu2023vgdiffzero,tian2023diffuse,ma2023diffusionseg} that highlight the potential of attention blocks to group pixels. 
Realizing such a capability with a unified representation for image and entity-level segmentation masks could potentially refine image generation, achieving greater coherence between the synthesized entities and their masks. 
Moreover, this unified representation offers significant potential for performing various dense prediction tasks, including both generation and segmentation in a single representation, as shown in Figure~\ref{fig.teaser}.

The intuitive solution is to represent segmentation masks simply as a colormap like Painter~\cite{wang2023images} and InstructDiffusion~\cite{geng2023instructdiffusion}.
However, the implementation is far from straightforward for three main reasons. 
First, the colormap design should be consistent with latent space not explored in Painter~\cite{wang2023images}.
Second, it should be able to differentiate entities in the same category. This challenge is not addressed in instructDiffusion~\cite{geng2023instructdiffusion}, which can only detect one entity. 
Finally, the mask quality is not guaranteed and usually has many noises without regular segmentation loss functions like cross-entropy or dice loss. 
Even though the colormap design effectively achieves a unified representation, the large-scale dataset requirements for training diffusion models are at odds with the sparse segmentation annotations at hand, resulting in a critical bottleneck in our exploration.

To tackle these challenges, our first step is to validate that variational autoencoder (VAE)~\cite{kingma2019introduction} used in stable diffusion~\cite{rombach2022high} can effectively encode and decode colormaps in the same way as images. 
Based on colormap representation and latent diffusion model, we introduce the UniGS framework to simultaneously generate images and multi-class entity-level segmentation masks. 
The UniGS has a UNet architecture augmented with dual branches: one for image and another for mask generation. 
In the mask branch, we propose two modules, including a location-aware palette and a progressive dichotomy module. The former assigns each entity area to some fixed colors by the entities' center-of-mass location, enabling UniGS to discriminate entities within the same category. The latter efficiently decodes generated noisy colormap into explicit masks without knowing the entity numbers. 

Then, we train our diffusion model under the inpainting protocol, addressing the scarcity of large-scale mask annotations. In this way, the diffusion model is primed to hone in on specific regions rather than the entire image. This flexibility facilitates using multiple segmentation datasets for training our diffusion model.
Combining unified image and mask representation with an inpainting pipeline further integrates various tasks within a single representation with minor modifications. Figure~\ref{fig.teaser} shows the effectiveness of the UniGS on four tasks, including multi-class multi-region inpainting, image synthesis, referring segmentation, and entity segmentation. 

The main contributions of this work are as follows:
\begin{itemize}
\item We are the first to propose a unified diffusion model (\ourmethod) for image generation and segmentation within a unified representation by treating the entity-level segmentation masks the same as images.

\item Two novel modules, including a location-aware palette and progressive dichotomy module, can make efficient transformations between the entity-level segmentation masks and colormap representations.

\item The inpainting-based protocol addresses the scarcity of large-scale segmentation data and affords the versatility to employ a unified representation across multiple tasks.

\item The extensive experiments show our UniGS framework's effectiveness on image generation and segmentation. In particular, UniGS can obtain segmentation performance comparable to state-of-the-art methods without any standard segmentation loss design. Our work can inspire foundation models with a unified representation for two mainstream dense prediction tasks.
\end{itemize}

\section{Related Work}
\label{sec:related_work}

\noindent
\textbf{Diffusion Model for Generation.} The diffusion models were initially introduced in the context of generation tasks~\cite{dhariwal2021diffusion} and have undergone significant evolution through latent design~\cite{rombach2022high}. Diffusion models have been applied to a wide variety of generation~\cite{rombach2022high,nichol2021glide,sinha2021d2c,dockhorn2021score}, image super-resolution~\cite{batzolis2021conditional,chung2022come,wu2023hsr}, image inpainting~\cite{lugmayr2022repaint,xie2023smartbrush,svitov2023dinar,yang2023paint}, image editing~\cite{kawar2023imagic,zhang2023sine,wu2023latent}, image-to-image translation~\cite{tumanyan2023plug,li2023bbdm,cheng2023general,zhang2023adding}, among others. We note that all current methods utilize the latent code to generate high-resolution images and have been extended to 3D~\cite{xu2023geometric,koo2023salad,metzer2023latent} or video generation~\cite{ho2022imagen,wu2023tune,luo2023videofusion,ceylan2023pix2video}. Instead of those methods focusing on content generation, we endow the diffusion model with perception and segmentation ability by using similar representations for the images.
\begin{figure*}[!tbp]
\centering
\includegraphics[width=0.85\textwidth]{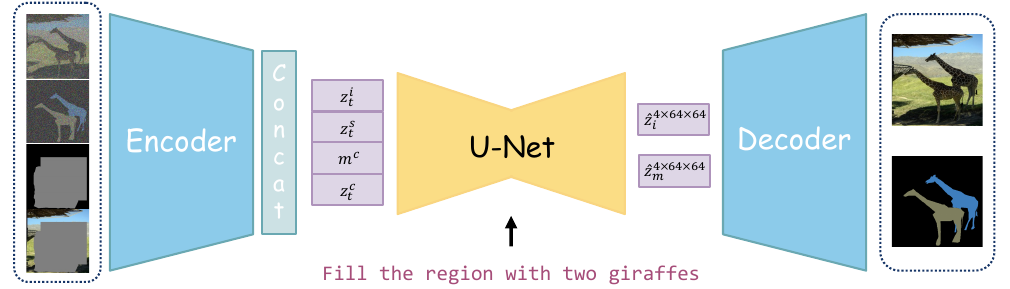}
\caption{\textbf{Overview of the UniGS framework within the inpainting pipeline.} Similar to stable diffusion, our UniGS denoise the feature in the latent space by an encoder and decoder. We note that the predictions of UNet $\Hat{z}_i$ and $\Hat{z}_m$ are unified representations that can be decoded into images and colormaps by a similar latent decoder. }
\label{fig.framework}
\end{figure*}
\vspace{1mm}

\noindent
\textbf{Diffusion Model for Segmentation.} Several studies have delved into pixel-level segmentation~\cite{chen2017deeplab,qi2021multi,zhao2017pyramid,qi2021pointins,he2017mask,qi2023aims,shen2022high,liu2018path,qi2022open,qi2023high} using diffusion models through three distinct pipelines. The first two pipelines emphasize leveraging pre-trained stable diffusion~\cite{rombach2022high} to simultaneously generate segmentation masks and images. Specifically, the first pipeline, as discussed in studies like~\cite{chefer2023attend,burgert2022peekaboo,liu2023vgdiffzero,tian2023diffuse,ma2023diffusionseg}, employs both self- and cross-attention maps in stable diffusion for shape grouping. However, these approaches demonstrate limited capabilities in the instance or entity-level discrimination~\cite{qi2023high,qi2022open}. Conversely, the second pipeline~\cite{wu2023datasetdm,wu2023diffumask,li2023guiding,xu2023open} primarily focuses on integrating a segmentation branch to produce precise mask generation by bringing substantially computational costs. Instead, the third pipeline~\cite{chen2023generalist,chen2023diffusiondet} is conditioned upon the input image by diffusing the image features to masks or bounding boxes. Furthermore, a prevalent issue with these methods is the inconsistency between image generation and segmentation mask generation processes. In contrast, we develop a unified representation for both tasks by converting the segmentation mask into a colormap.

\vspace{1mm}
\noindent
\textbf{Unified Representation.} Some foundation models~\cite{wang2023images,wang2023seggpt,geng2023instructdiffusion} explored unified representation for both generation and perception tasks. Our work is mostly similar to the Painter~\cite{wang2023images} and InstructDiffusion~\cite{geng2023instructdiffusion} but with various designs. Rather than reproducing the original color through MAE's~\cite{he2022masked} regression as in Painter~\cite{wang2023images}, our approach involves gradually diffusing the latent code of the colormap by several time steps. Compared to the InstructDiffusion~\cite{geng2023instructdiffusion}, our framework offers greater flexibility in decoupling the image and colormap using distinct latent codes. As a result, there's no necessity to employ a lightweight segmentation branch for mask generation in our approach.
\section{Method}
Based on the latent diffusion model~\cite{rombach2022high}, the proposed UniGS framework aims to progressively and simultaneously denoise images and segmentation masks given a text prompt. In Figure~\ref{fig.framework}, we show the overview of the UniGS model within the inpainting pipeline. Such a pipeline can address the challenge of insufficient segmentation datasets and unifying multiple tasks in a single representation.

Specifically, the input of our UNet has four parts, including the latent encode of the noised image, colormap, context, and a resized coarse mask. They are denoted by $z_t^i$, $z_t^s$, $z_t^c$ and $m^c$, respectively. Based on the text prompt, we use an UNet to denoise the $z_t^i$, $z_t^s$ to $\Hat{z}_i$ and $\Hat{z}_m$. During inference, $z_t^i$ and $z_t^s$ would be the pure Gaussian noise. Compared to stable diffusion~\cite{rombach2022high}, there is no obvious structure difference except for the input and output channel numbers.

\begin{table}[tbp!]
\centering
\small
\setlength\tabcolsep{3pt}
\begin{tabular}{cccc}
\toprule
Notation & Definition & Notation & Definition  \\ \midrule
$\Upsilon$ & AutoEncoder (VAE) & $\Omega$ & Coarse Mask Generator \\
$\Psi$ & Colormap Encoder & $\Phi$ & Colormap Decoder  \\ \midrule
$M$ & Entity-level Masks & $M_c$ & Colormap \\
$I_0$ & Original Image & $m_c$ & Coarse Mask\\ 
\bottomrule
\end{tabular}
\caption{\textbf{Illustration of some notations in the Method section.}}
\label{tab:notation}
\vspace{-4mm}
\end{table}

In the following, we begin with an overview of latent diffusion techniques for high-resolution image synthesis. Then, we introduce our novel mask representation to represent entity masks. Lastly, we propose our whole inpainting pipeline and its extension to multiple tasks. It is noted that Table~\ref{tab:notation} lists essential notions in this section. 

\subsection{Review of Latent Diffusion} 
Diffusion models~\cite{ho2020denoising} is a class of likelihood-based models that define a Markov chain of forward and backward processes, gradually adding and removing noise to sample data. The forward process is defined as
\begin{equation}
\label{eq:noise_process}
    q(z_t | z_0) = \mathcal{N}(z_t | \sqrt{\bar{\alpha}_t} z_0, (1 - \bar{\alpha}_t) z_0),
\end{equation}
which transforms data sample $z_0$ to a latent noisy sample $z_t$ for $t\in\{0, 1, ...,T\}$ by adding gausian noise $\epsilon$ to $z_0$.
$\bar{\alpha}_t \coloneqq \prod_{s=0}^{t} \alpha_s = \prod_{s=0}^{t} (1 - \beta_s)$ where $\beta_s$ represents the noise variance schedule~\cite{ho2020denoising}.
During training, a neural network (usually an UNet) $f_\theta(\bm{z}_t, t)$ is trained to predict $\epsilon$ to recover $z_0$ from $z_t$ by minimizing the training objective with $\ell_2$ loss~\cite{ho2020denoising}:
\begin{equation}
\label{eq:diff_loss}
\mathcal{L}_\text{train} =  \frac{1}{2}|| f_\theta(z_t, t) - \epsilon ||^2.
\end{equation}
where $\theta$ is the parameters of the neural network. At inference stage, data sample $z_0$ is reconstructed from $z_T$ with the model $f_\theta$ and an updating rule~\cite{ho2020denoising, song2021denoising} in an iterative way, \ie,  $z_T \rightarrow z_{T-\Delta} \rightarrow ... \rightarrow z_0$. For a clear illustration, we omit the updating rule and regard the output of $f_\theta(z_t, t)$ as $z_0$.
In the context of generating high-resolution images $I_0 \in \mathcal{R}^{h\times w \times 3}$ given a text prompt, diffusion models would incur substantial computational costs if using large image size like $h=w=512$. The $h$ and $w$ represent the image height and width. To tackle this issue, latent diffusion models (LDM) uses latent code of the images as $z_0^i \in \mathcal{R}^{\frac{h}{4}\times \frac{w}{4} \times 4}$~\cite{van2017neural}:
\begin{equation}
z_0^i = \Upsilon(I) \quad \text{and} \quad \Hat{I}_{0} = \Upsilon^{-1}(\Hat{z}_{0}^{i})
\end{equation}
where $\Upsilon$ and $\Upsilon^{-1}$ represent the encoding and decoding process of AutoEncoder (VAE) to $I$. $\Hat{*}$ indicates the prediction results. 
As such, latent diffusion reduces computational demands and maintains good generation ability. We base our UniGS model on Stable Diffusion~\cite{rombach2022high}, a popular LDM variant.

\subsection{Colormap-based Entity Mask Representation}
\begin{figure}[!tbp]
\centering
\includegraphics[width=0.47\textwidth]{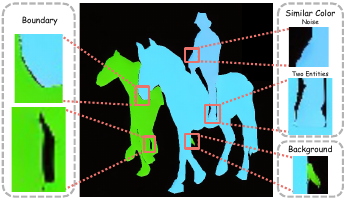}
\caption{\textbf{Illustration of several difficult cases in the decoded color map.} We conclude those cases into three kinds of problems, including boundary, similar color, and background.}
\label{fig.color_noise}
\vspace{-2mm}
\end{figure}

We use a colormap representing segmentation masks that can align mask representation with image format while supporting variability in entity numbers. However, designing a colormap encoder and decoder is non-trivial due to the requirements of discriminating each entity within the same categories. Moreover, this representation would lack the standard segmentation loss in latent space like binary cross-entropy and dice loss. Using the denoise loss in Eq~\ref{eq:diff_loss} for colormap would lead to several extreme cases in Figure~\ref{fig.color_noise}. 
Thus, we describe our location-aware palette and progressive dichotomy modules in colormap encoding and decoding to solve the above-mentioned problems.

\vspace{1mm}
\noindent \textbf{Colormap Encoding.} 
The colormap encoder $\Psi$ converts several entity-level binary segmentation masks $M \in \{0, 1\}^{n\times h\times w}$ to an colormap $M_c \in [0, 255]^{h\times w\times 3}$ as
\begin{equation}
M_c = \Psi(M) 
\end{equation}
$n$ denotes the number of sampled entities. The $M_c$ is initialized by zero value and then assigned some color for each entity area by our location-aware palette. Specifically, we partition an image into $b \times b$ grids where each grid has a unique color. Each entity area is associated with these fixed colors if their gravity centers are at the grids. Each RGB channel has five candidate color values $\{0,64,128,192,255\}$ in our location-aware palette. Thus, the overall color number is $124=5^3-1$ with color $(0, 0, 0)$ indicating the background. The grid numbers $b^2=|b\times b|$ should be less than 124. 

The location-aware palette design proves simple but efficient in covering nearly all labeled entities (97.4\% coverage ratio across the COCO, ADE20K, OpenImages, and EntitySeg datasets). That's because UNet has a position encoding design that can help predict corresponding colors. In contrast, random color assignments often struggle to distinguish between entities of the same category due to providing too large a color space.

\vspace{1mm}
\noindent \textbf{Colormap Decoding.}
While the generated colormap effectively differentiates between entities visually, converting it into the perfect entity-level masks presents several challenges. A primary issue is the need for more awareness of entity numbers. Therefore, heuristic k-means clustering is impractical. To tackle this issue, we propose a progressive dichotomy module $\Phi$ to group areas of identical color by pixel-level features $p$ without prior knowledge of cluster numbers.

\begin{equation}
\hat{M} = \Phi(\hat{M_c}) = \Phi(\Upsilon^{-1}(\hat{z}_0^s))
\end{equation}
where $\hat{M_c}$ is predicted colormap decoded by VAE, and  $\hat{M} \in \{0,1\}^{n\times H\times W}$ has $n$ binary masks. 

Specifically, the progressive dichotomy module (PDM) is a depth-first cascaded clustering method where we further split the $j^{\text{th}}$ entity mask $\hat{m}_{v-1}^{j}$ generated at ${(v-1)}^{\text{th}}$ iteration into the two sub-masks $\hat{m}_{v}^{2j}$ and $\hat{m}_{v}^{2j+1}$ at ${v}^{\text{th}}$ iteration, 
\begin{equation}
\{\hat{m}_{v}^{2j}, \hat{m}_{v}^{2j+1}\} = \mathcal{BK}(\hat{m}_{v-1}^{j})
\end{equation}
The $\mathcal{BK}$ denotes two-cluster k-means and each $\hat{m} \in \{0, 1\}^{H\times W}$ . Further splitting $\hat{m}_{v-1}^{j}$ will stop until the average L2 distance of mask pixels to their mean less than $\delta$:
\begin{equation}
\frac{\sum_{o\in \hat{m}_{v-1}^{j}} (p_o-c_{\hat{m}_{v-1}^{j}})^{2}}{|\hat{m}_{v-1}^{j}|} < \delta
\end{equation}
The $c_{\hat{m}_{v-1}^{j}} = \frac{\sum_{o\in \hat{m}_{v-1}^{j}} p_o}{|\hat{m}_{v-1}^{j}|}$ with $|\hat{m}_{v-1}^{j}|$ denoting the pixel numbers of mask $\hat{m}_{v-1}^{j}$.

The pixel feature $p_o$ is designed in light of three critical observations in Figure~\ref{fig.color_noise}. $o \in [0,...,h\times w)$. At first, it is not trivial to discern whether a gradient color signifies one or multiple entities. Second, the foreground colors would be degraded by the background. Thirdly, some black holes are hard to predict as true or false positives. Thus, we design $p_o \in \mathcal{R}^{1\times6}$ with both RGB and LAB image space. Including LAB image space is pivotal due to their perceptual uniformity property, which ensures that minor variations in LAB values translate to approximately uniform alterations in color as perceived by the human eye, thereby providing enhanced contrast.

\begin{table*}[tp]
\centering
\small 
\setlength\tabcolsep{1pt}
\begin{tabular}{cccccc}
\toprule
\multirow{2}{*}{Task} & 
\multicolumn{3}{c}{Condition} &
\multicolumn{2}{c}{Output} \\ \cline{2-6}
& coarse mask ($m^c$) & control factor ($z_i^c$) & text prompt template & image ($\hat{z}^i_0$) & mask ($\hat{z}^s_0$) \\ \toprule
Inpainting & $\Omega(M)$ & $\Upsilon(I_0 \odot (1-m^c))$ & `inpainting: generate dog.' & \cmark & \cmark \\ 
Image Synthesis & $\mathcal{J}$ & $\Upsilon(M_c)$ & `synthesis: generate dog, ground, and sky.' & \cmark & \xmark \\ 
Referring Segmentation & $\Omega(M)$ & $\Upsilon(I_0 )$ & `referring: find dog.' & \xmark & \cmark \\ 
Entity Segmentation & $\mathcal{J}$ & $\Upsilon(I_0)$ & `panoptic: all entities.'& \xmark & \cmark \\ \bottomrule
\end{tabular}
\caption{\textbf{Illustration of the condition signal's design on training task in our framework.} The \cmark and \xmark indicate whether we expect the two output tensors to be the same as our condition $z_i^c$.}
\label{tab:multi_task_learning}
\vspace{-2mm}
\end{table*}

\subsection{Inpainting Pipeline}
\label{sec:inpaint}
We adopt an inpainting pipeline for training and inference to reconcile the generative model's requirements for large-scale segmentation datasets. For example, the Open-Images dataset~\cite{benenson2019large} with mask annotations encompasses approximately 1.8 million images but only contains about three entity-level labels per image. Directly training the latent diffusion model results in too many ambiguities due to unlabeled areas. Instead, our inpainting pipeline enables the generative model to concentrate on the valid areas regardless of the partial segmentation labels.

In the training period, the UNet input is $z^{u} \in \mathcal{R}^{\frac{H}{4}\times \frac{W}{4}\times 13}$ that concatenated by
\begin{equation}
z_t^u = \text{CONCAT}(z_t^i, z_t^s, m^c, z_t^c)
\end{equation}
$z_t^i$ and $z_t^s$ are the latent code of the noised image $I_t$ and colormap $S_t$ in time step $t$ where both $z_t^i$ and $z_t^s$ are $\mathcal{R}^{\frac{h}{4}\times \frac{w}{4}\times 4}$. 
$m^c \in\{0,1\}^{\frac{h}{4}\times \frac{w}{4}\times 1}$ is a coarse mask where $1$ indicates a rectangular or an irregular area that needs our UniGS framework to fill entities and their masks, 
\begin{equation}
m_c = \Omega(M)
\end{equation}
More details regarding  $\Omega$ are available in the supplementary material. 
Next, $z_i^c$ is the latent code of the masked image by $m_c$, 
\begin{equation}
z_t^c = \Upsilon(I_0 \odot (1-m^c))
\end{equation}
The UNet output is 
\begin{equation}
\hat{z}_0 = f_\theta(z_t^u, t) \in \mathcal{R}^{\frac{H}{4}\times \frac{W}{4}\times 8}
\end{equation}
where the first and last four channels of $\hat{z}_0$ can be latently decoded to the final image and colormaps.

\subsection{One-to-Many Tasks}
The inpainting pipeline with colormap representation allows for integrating various tasks within a single model. 
We use the UniGS model for four vision tasks: inpainting, image synthesis, referring segmentation, and entity segmentation. The configuration of each task is presented in Table~\ref{tab:multi_task_learning}.

\vspace{1mm}
\noindent \textbf{Multi-class Multi-region Inpainting.} Our baseline task that has been detailed in Section~\ref{sec:inpaint}.

\vspace{1mm}
\noindent \textbf{Image Synthesis:} $z_i^c$ is latent code of colormap $M_c$ containing sampled entities. Meanwhile, the coarse mask $m_c$ is $\mathcal{J}$ matrix of all ones to cover the entire image area. For the output, we maintain $\hat{z}^s_0=\Upsilon(M_c)$ and predict the entities' appearance $\hat{z}^i_0$.

\vspace{1mm}
\noindent \textbf{Referring Segmentation.} This task aims at segmenting some classes based on instructions. Thus, we preserve image information by $z_i^c=\Upsilon(I_0)$. Considering the requirement of negative samples to ensure alignment between the entity and text prompt, we define $\lambda$ as the possibility of each sampled entity belonging to a negative in training. For negative samples, the category names in text prompts are replaced with others that do not appear in the coarse mask.

\vspace{1mm}
\noindent \textbf{Entity Segmentation.} All the entities should be predicted in $\hat{z}^s_0$ with the coarse mask area $\mathcal{J}$.

\begin{figure*}[tp]
    \centering
    \includegraphics[width=0.99\textwidth]{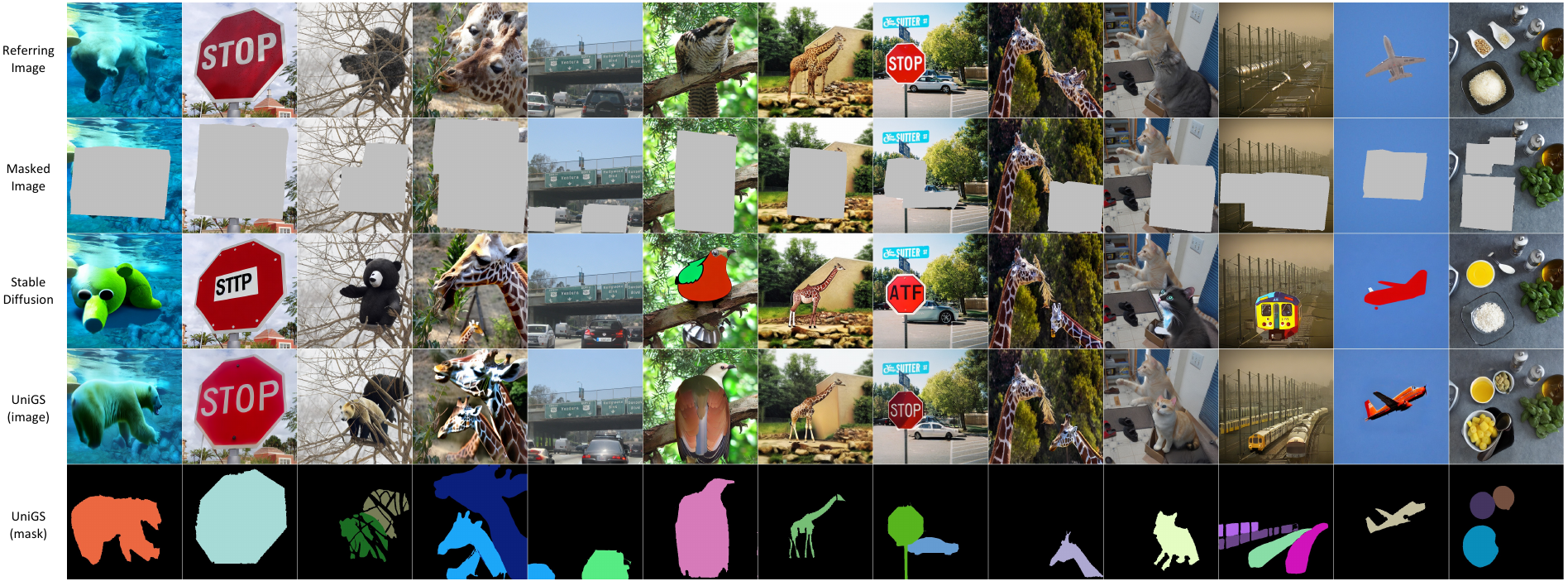}
    \caption{\textbf{Qualitative comparison of inpainting results between Stable Diffusion and our UniGS.} For the coarse masks, we keep the consistency of the ones used in our training phase to eliminate the pattern gap. Furthermore, we showcase multi-class, multi-region inpainting to the multiple entities within the same category, moving beyond the conventional approach of incorporating a single entity.}
    \label{fig.mask2img_visual}
    \vspace{-2mm}
\end{figure*}

\section{Experiments}
In this section, we first explore the performance of our proposed UniGS in four individual tasks, including multi-class multi-region inpainting, image synthesis, referring segmentation, and entity segmentation. 
Some key module designs or hyper-parameters on the mask quality are ablated in referring segmentation.
Similar to other works~\cite{chen2023anydoor,ruiz2023dreambooth,qi2023high} to evaluate the image and mask quality, we use the intersection over union (IoU) and recall for mask evaluation and the Fréchet inception distance (FID) and CLIP score (CS) for image generation.

\subsection{Experiment Setting}
For each single-task model, we exclusively utilize the COCO dataset~\cite{lin2014microsoft}, the Open Images~\cite{OpenImages}, and EntitySeg datasets~\cite{qi2023high} as our training data. Considering the COCO panoptic data having about 10$\%$ ignored area, we only use the EntitySeg for entity segmentation task in case of performance degradation.

In our training process, we randomly sample up to four objects per sampled area for tasks such as inpainting, image synthesis, and referring segmentation. On the other hand, entity segmentation should include all the entities that can cover the whole sampled area. During the inference period, we sample 1000 images in COCO validation data as our test set, where each image has a coarse mask and various control factors for different tasks, as shown in Table~\ref{tab:multi_task_learning}.

We initialize our model with stable diffusion v1.5 inpainting and weight newly added channels as zero. The image size and latent factor reduction ratio are set to 512$\times$512. 

\subsection{Multi-class Multi-region Inpainting}
\begin{table}[tp]
  \centering
  \small
  \setlength\tabcolsep{3pt}
  \begin{tabular}{@{}ccccc@{}}
    \toprule
    \multirow{2}*{Method} & FID ($\downarrow$) & CLIP Score ($\uparrow$) & FID ($\downarrow$) & CLIP Score ($\uparrow$)\\
    \cmidrule(lr){2-3} \cmidrule(lr){4-5} 
& \multicolumn{2}{c}{single object} & \multicolumn{2}{c}{multiple objects} \\
    \midrule
    SD$_{\text{I}}^{\text{1.5}}$ &  4.95 & 88.86 & 7.82 & 83.80 \\
    UniGS$^{\text{1.4}}$  & 4.39 & 88.92 & 6.19 & 84.43 \\
    UniGS$_{\text{I}}^{\text{1.5}}$ & \textbf{3.78} & \textbf{90.22} & \textbf{5.89}  & \textbf{85.87} \\
    \bottomrule
  \end{tabular}
  \caption{\textbf{Quantitative results on image inpainting task.} The SD$_{\text{I}}^{\text{1.5}}$ means the stable diffusion inpainting model with version 1.5. The UniGS$^{\text{1.4}}$ is the UniGS that initialized from the stable diffusion model with version 1.4, the UniGS$_{\text{I}}^{\text{1.5}}$ is the UniGS that initialized from stable diffusion inpainting model with version 1.5.}
  \label{tab:mask2img_compare}
  \vspace{-2mm}
\end{table}

We evaluate the inpainting model by inserting one or multiple objects into the coarse mask area generated from the entity masks. 
The model's output in this task includes the generated image and colormap. 
As in Table \ref{tab:mask2img_compare}, our method outperformed the original model regarding both FID and CLIP scores. 
This improvement was observed even when our model was initialized using the stable diffusion 1.4 pre-trained model, highlighting the effectiveness of our approach in enhancing image generation through the integration of object mask guidance.
That's because our unified representation effectively constrains the model to maintain consistency between the visual appearance of the objects and their corresponding masks. 
As a result, the object masks impart a robust shape priority, guiding and refining the image generation process to ensure alignment and coherence in the final output.

Figure~\ref{fig.mask2img_visual} shows the visual comparison between stable diffusion and our UniGS with the coarse masks generated from our code used in the training period. In other words, our testing keeps a similar pattern of inpainting area. It is evident from these results that the objects generated by the model are in strong harmony with the high-quality masks, showcasing the model's effectiveness in seamlessly integrating the objects into the overall image composition. Furthermore, this impressive coherence between generated objects and their masks is attributable to our model's unified representation of images and segmentation masks.

\begin{figure*}[tp]
    \centering
    \includegraphics[width=0.99\textwidth]{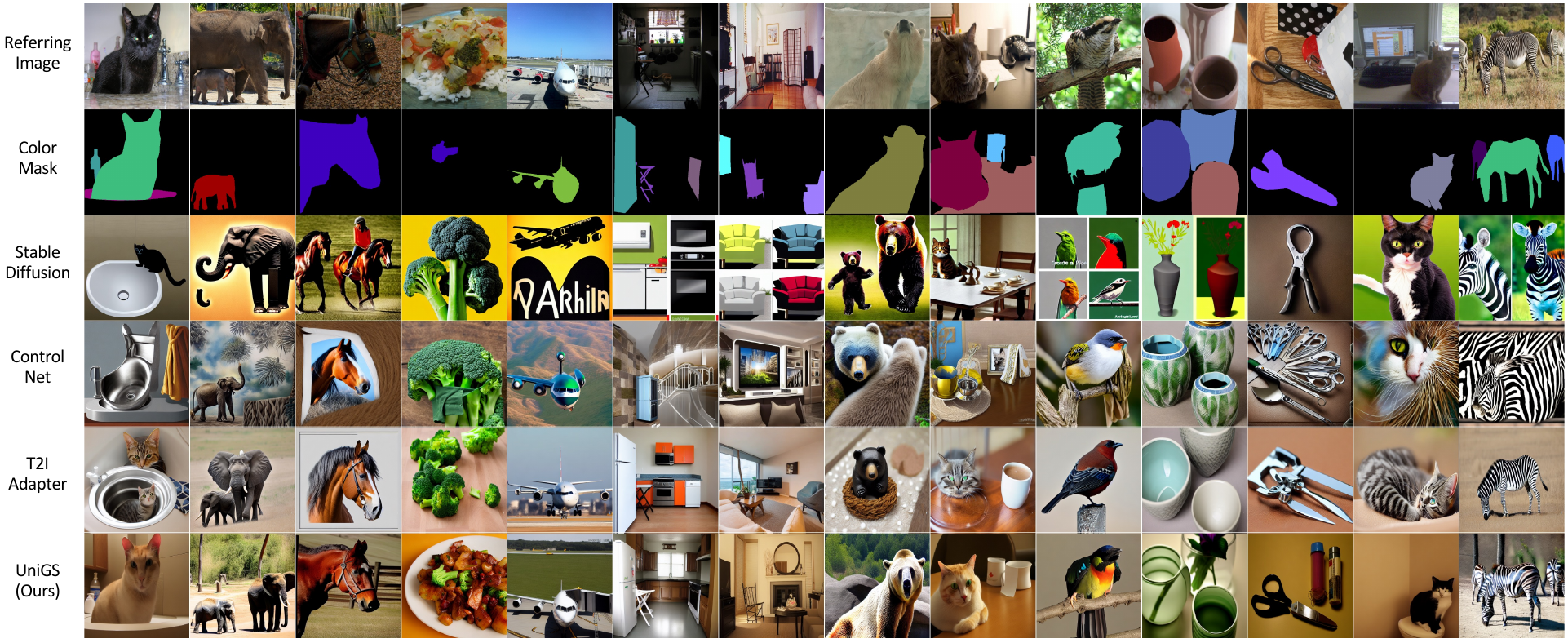}
    \vspace{-3mm}
    \caption{\textbf{Qualitative comparison among Stable Diffusion, ControlNet, T2I-Adapter, and our UniGS on the image synthesis task.} Compared to those methods, UniGS maintains more coherence between the generated entities and their corresponding segmentation masks.}
    \label{fig.image_synthesis_visual}
    \vspace{-4mm}
\end{figure*}

\begin{figure*}[!tp]
\centering
\includegraphics[width=0.99\textwidth]{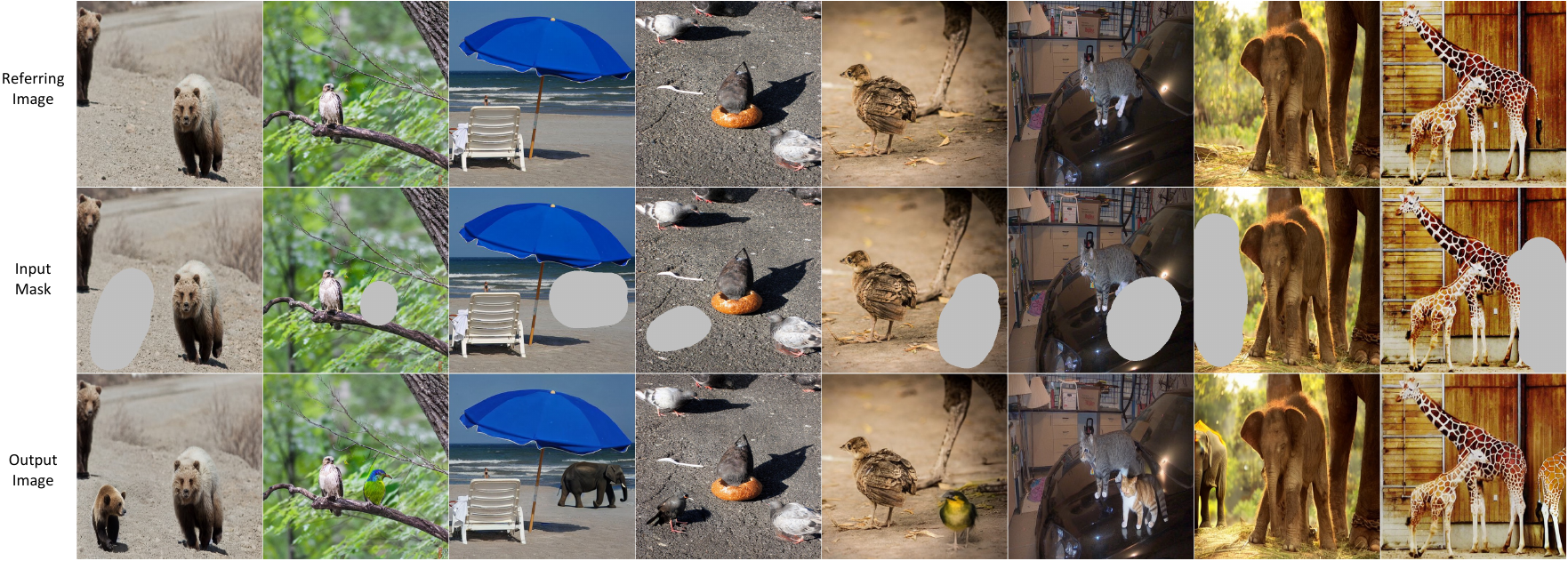}
\vspace{-3mm}
\caption{\textbf{Visualization results of our UniGS in the real world.} We generate the coarse masks used in inpainting by brush strokes from Gradio. We identified some interesting observations, particularly the appearance of shadows from the third to sixth columns.}
\vspace{-4mm}
\label{fig.object_insert_visual}
\end{figure*}

Figure~\ref{fig.object_insert_visual} presents the visualization results with real-world irregular coarse masks generated from brush strokes. 
Our UniGS model still has considerable ability in object insertion without the domain gap problem. 
Also, some exciting things can be observed. 
For example, our model can generate realistic shadows of the inserted objects in the last two columns, which are rarely observed in other generation models. 
This capability emerges despite the absence of explicit shadow supervision during the training phase. 
The reason might be that the target of mask generation forces our UniGS model to learn the grouping behavior of pixels with similar textures. 
Thus, it is easier for our UniGS to recognize and replicate shadow patterns from their context that has shadows. 
This attribute enhances the overall realism and visual coherence of the inserted objects.
%
\subsection{Image Synthesis}
In this task, we expect to take a colormap as input along with a text-based image synthesis prompt and output a synthesized image. 
Except for the conventional metrics of Fréchet inception distance (FID) and CLIP score, we incorporated mean Intersection over Union (mIoU) to evaluate the alignment of the generated image with the specified mask shape.
For this external evaluation, we utilized the Mask2Former model equipped with a large swan backbone to perform segmentation on the images generated by our model. 
The mIoU is then calculated by comparing these segmentation masks against the original colormap.
In Figure~\ref{fig.image_synthesis_visual}, we present the visual consistency between the synthesized objects and the provided color masks among four methods: stable diffusion, ControlNet, T2I Adaptor, and UniGS.
We modify the pipeline of the compared approaches for a fair comparison.
For stable diffusion designed not for image synthesis, we only use text prompts for conditions.
For ControlNet and T2I-Adaptor, we follow the default settings and input the segmentation map and text prompt to get the synthesis image. 
In this figure, we highlight the model's ability to align the generated objects with the specified mask constraints closely.
Moreover, those visualization results reflect the successful and seamless integration of these objects within their backgrounds, further highlighting the benefits of our unified representation in synthesizing contextually coherent and visually harmonious images.
As shown in Table~\ref{tab:img_syntheis_compare}, our method is more favorable in all critical metrics, including the FID, CLIP score, and mIoU. This comparison underscores that the unified representation for both image and segmentation mask can help the image synthesis network have a higher quality perceptual judgment and more precise mask-to-object alignment.


\begin{table}[tp]
  \centering
  \small
  \setlength\tabcolsep{3pt}
  \begin{tabular}{@{}lcccccc@{}}
    \toprule
    \multirow{3}*{\textbf{Method}} & FID ($\downarrow$) & CS ($\uparrow$) & mIoU ($\uparrow$) & FID ($\downarrow$) & CS ($\uparrow$) & mIoU ($\uparrow$) \\
    \cmidrule(lr){2-4} \cmidrule(lr){5-7} 
    & \multicolumn{3}{c}{single object} & \multicolumn{3}{c}{multiple object} \\
    \midrule
    SD~\cite{xu2023geometric} & 36.502 & 54.708 & 0.196 & 34.770 & 56.511 & 0.191 \\
    CN~\cite{zhang2023adding} & 35.111 & 55.230 & 0.277 & 30.108 & 58.709 & 0.326 \\
    T2I~\cite{mou2023t2i} & 34.434 & 59.024 & 0.306 & 24.898 & 62.910 & 0.379 \\
    \midrule
    \textbf{\ourmethod} & \textbf{15.272} & \textbf{65.015} & \textbf{0.781} & \textbf{14.271} & \textbf{69.504} & \textbf{0.777} \\
    \bottomrule
  \end{tabular}
  \caption{\textbf{Quantitative results on image synthesis task.} `SD', `CN', and T2I indicate the stable diffusion, CrontrolNet, and T2I-Adaptor. The `CS' is the CLIP Score.}
  \label{tab:img_syntheis_compare}
  \vspace{-5mm}
\end{table}

\subsection{Referring Segmentation}
We evaluate the quality of generated masks by mIoU and recall metrics for the referring segmentation task.
In Table~\ref{tab:img2mask_compare} on the comparison to the state-of-the-art segmentation method Mask2Former \cite{Cheng_2022_CVPR}, our generative method has considerable segmentation quality.
These results are worth noticing as we do not use any explicit segmentation loss, thereby demonstrating the potential of the UniGS model. 

\begin{table}[!tbp]
  \centering
  \small
  \begin{tabular}{@{}lccc@{}}
    \toprule
    Method & Backbone & mIoU($\uparrow$) & Recall ($\uparrow$) \\
    \midrule
    Mask2Former~\cite{Cheng_2022_CVPR} & Swin-Large & \textbf{0.815} & \textbf{0.887} \\
    \ourmethod \ (Ours) & - & 0.808 & 0.872 \\

    \bottomrule
  \end{tabular}
  \caption{\textbf{Quantitative results on referring segmentation task.} We choose Mask2Former with a Swin-Large backbone as our baseline for comparison with SOTA segmentation methods.}
  \label{tab:img2mask_compare}
  \vspace{-3mm}
\end{table}


\begin{table}[tp]
  \centering
  \small
  \setlength{\tabcolsep}{2pt}
  \begin{tabular}{@{}lcccc@{}}
    \toprule
    Method & Backbone & mIoU($\uparrow$) & Recall ($\uparrow$) & AP$^e$ ($\uparrow$)\\
    \midrule
    ConInst-Entity~\cite{tian2020conditional,qi2022open} & Swin-Large & 0.621 & 0.685 & 0.397  \\
    SAM~\cite{Kirillov_2023_ICCV} & VIT-Huge & 0.653 & 0.714 & 0.432 \\
    CropFormer~\cite{qi2023high} & Swin-Large & \textbf{0.664} & \textbf{0.727} & \textbf{0.449}\\
    \ourmethod \ (Ours) & - & 0.631 & 0.692 & 0.407 \\

    \bottomrule
  \end{tabular}
  \caption{\textbf{Quantitative results on entity segmentation task}. The AP$^{e}$ is AP  with a non-overlapped constraint used in entity segmentation.}
  \label{tab:entityseg_compare}
  \vspace{-6mm}
\end{table}
\subsection{Entity Segmentation}
%
The entity segmentation aims at splitting an input image into several semantically coherent regions. 
Thus, the generated colormap should cover the whole image.
After latent decoding the output from UNet, we use the progressive dichotomy module to transform the colormap into explicit segmentation masks.
In Table~\ref{tab:entityseg_compare}, we show that there is still a significant performance gap between the UniGS and state-of-the-art entity segmentation model.
However, we mention that the entity performance of the UniGS model is acceptable and better than kernel-based methods like CondInst.

\begin{table}[tp]
\centering
\small
\begin{tabular}{@{}lcc@{}}
\toprule
Method & mIoU($\uparrow$) & Recall ($\uparrow$) \\
\midrule
Random Color Assignment & 0.493 & 0.563 \\
Location-aware Palette (Ours) & \textbf{0.808} & \textbf{0.872} \\
\bottomrule
\end{tabular}
\caption{\textbf{Ablation study on various color assignments in mask encoder.} The `Random Color Assignment' indicates assigning each entity with a random color.}
\vspace{-6mm}
\label{tab:color_map}
\end{table}

\subsection{Ablation Study}
In the following, we ablate different color assignment criteria and progressive dichotomy modules with various hyper-parameters.
All the ablation studies are conducted in referring segmentation tasks to measure the mask quality.

\vspace{1mm}
\noindent \textbf{Location-aware Palette.}
To evaluate the effectiveness of our color mapping over the random color assignment for the object, we have individually trained the referring segmentation models for both color mapping methods, as shown in Table \ref{tab:color_map}. Our color mapping method lets the model easily learn the pattern of the object color mask.

\vspace{1mm}
\noindent \textbf{Progressive Dichotomy Module.} Compared to the fixed cluster numbers in k-means, our proposed progressive dichotomy module has the advantage of adaptive cluster numbers. 
We verify our method in Table~\ref{tab:kmeans_method_compare} by comparing K-Means and ours.
Our progressive dichotomy module has no noticeable performance degradation compared to K-Means, even with knowing the ground truth numbers, manifesting the effectiveness and robustness of our progressive dichotomy module. 

Furthermore, the distance threshold $\delta$ and pixel feature used in the progressive dichotomy module are ablated in
Table~\ref{tab:dist_thresh}. In Table~\ref{tab:dist_thresh}(a), we can see that the distance threshold designed in the progressive dichotomy module is robust to the segmentation performance ranging from 0 to 20. In Table~\ref{tab:dist_thresh}(b), using the RGB and LAB space pixel feature to decode the generated colormap can obtain the best mask quality because the LAB space can offer more contrast information for those two similar colors in RGB space.

\begin{table}[tp]
  \centering
  \small
  \setlength{\tabcolsep}{2pt}
  \begin{tabular}{@{}lccc@{}}
    \toprule
    Method & Cluster Numbers& mIoU($\uparrow$) & Recall ($\uparrow$) \\
    \midrule
    \multirow{2}*{Native K-Means} & Fixed (3) & 0.520 & 0.641 \\
    & Adaptive (GT) & \textbf{0.810} & \textbf{0.874} \\ \midrule
    PDM & Adaptive & 0.808 & 0.872 \\
    \bottomrule
  \end{tabular}
  \caption{\textbf{Comparison between native K-Means and our progressive dichotomy module.} `Fixed (3)' indicates that we assign native K-Means with three cluster numbers. `Adaptive (GT)' is to assign the cluster number by the ground truth number.}
  \label{tab:kmeans_method_compare}
  \vspace{-2mm}
\end{table}

\begin{table}[t!]
\begin{minipage}{\textwidth}
\begin{minipage}[t]{0.22\textwidth}
        \centering
        \footnotesize
        \setlength{\tabcolsep}{2pt}
        \begin{tabular}{@{}ccc@{}}
        \toprule
        $\delta$ & mIoU($\uparrow$) & Recall ($\uparrow$) \\
        \midrule
        1 & 0.804 & 0.868 \\
        10 & \textbf{0.808} & \textbf{0.872} \\
        20 & 0.791 & 0.857 \\
        50 & 0.705 & 0.789 \\
        \bottomrule
      \end{tabular}
        
    (a)
        \end{minipage}
        \begin{minipage}[t]{0.24\textwidth}
        \centering
        \footnotesize
        \setlength{\tabcolsep}{2pt}
        \begin{tabular}{@{}ccc@{}}
        \toprule
        pixel feature & mIoU($\uparrow$) & Recall ($\uparrow$) \\
        \midrule
        RGB & 0.796 & 0.860 \\
        LAB & 0.787 & 0.856 \\
        RGB + LAB & \textbf{0.808} & \textbf{0.872} \\
        \bottomrule
      \end{tabular}
        
    (b)
    \end{minipage}
\end{minipage}
\vspace{-3mm}
\caption{\textbf{Ablation study on progressive dichotomy module.} We ablate the distance threshold $\delta$ (a) and pixel feature (b) in the colormap decoding process.}
\vspace{-6mm}
\label{tab:dist_thresh}
\end{table}

\section{Conclusion}
\vspace{-2mm}
This paper introduces a novel, effective unified representation in image generation and segmentation tasks. The key to our approach is regarding entity-level segmentation masks as a colormap generation problem. To distinguish entities within the same category, we employ a location-aware palette where each entity is distinctly colored based on its center-of-mass location. Furthermore, our progressive dichotomy module can efficiently transform a generated, albeit noisy, colormap into high-quality segmentation masks. Our extensive experiments on four diverse tasks demonstrate the robustness and versatility of our unified representation in image generation and segmentation. In the future, we will explore the multi-task training of our unified representation in a single model. We hope our work can foster the development of a foundation model with a unified representation for various tasks.

{
    \small
    \bibliographystyle{ieeenat_fullname}
    \bibliography{main}
}
\clearpage
\setcounter{page}{1}
\maketitlesupplementary
This supplementary material document provides more visualization results and training/inference details on Multi-class multi-region inpainting, image synthesis, referring segmentation, and entity segmentation. The supplementary material is organized as follows:

\begin{itemize}
\setlength{\itemsep}{1pt}
\setlength{\parsep}{1pt}
\setlength{\parskip}{1pt}
\item Training/inference details on four tasks.
\item More visualization results, including an example illustration of decoding colormap and the generation results of three tasks.
\end{itemize}

Also, please check \underline{\textit{a recorded video}} to obtain a brief description of our paper.

\section{Training/inference Details}
\paragraph{Multi-class multi-region inpainting}
In our research, we train our model on the COCO and Open-Images dataset, initializing it with the stable diffusion inpainting model v1.5. 
For each image, we sample a maximum of four objects and then.
Based on the ground truth mask, we have two schemes to make the coarse mask: 1. simulate a more coarse mask using the curve. 2. direct. The process follows in Algorithm~\ref{alg:code} that is used in Paint-by-Example.
This coarse mask is then cropped out from the original image. 
The cropped image and coarse mask are concatenated and fed into a UNet, with the expected output being an inpainted image and a separate color mask.
Under this setting, the model was trained for 48 epochs, resulting in the development of our model.
During inference, we sample a maximum of three objects. Similarly, we sample out coarse masks. 
And utilize randomly initialized noise for DDIM denoising, a total of 200 steps.

We have compared the number of parameters and the inference speed with the original stable diffusion inpainting model in Table~\ref{tab:param_supp}. 
We have only added a few channels to the input and output, keeping the parameter count almost consistent with that of stable diffusion. 
The inference speed also remains nearly the same. 
Without incurring additional computational costs, we have achieved better output results.

\paragraph{Image Synthesis} During image synthesis, we use a colormap as our conditional input instead of an image, and the input for the coarse mask is a mask entirely filled with ones, indicating that the area of interest is the entire image.
We conducted a full training over 48 epochs on the COCO and Open-Images dataset, with the maximum number of entity samples set to four. 
The inference process is consistent with the training but with a maximum sample number of three. 
The input includes a text condition, colormap, and an all-one coarse mask to produce the synthesized image.

\begin{table}[t!]
\centering
\small
\setlength{\tabcolsep}{2pt}
\begin{tabular}{@{}lcc@{}}
\toprule
Method & Parameters & Speed \\
\midrule
Stable Diffusion & 859.54 M & 14.48 \\
UniGS (Ours) & 859.56 M & 14.40 \\
\bottomrule
\end{tabular}
\caption{\textbf{Comparison of parameters and inference speed between Stable Diffusion and our UniGS.} The inference speed is tested by \textit{the seconds per image}. We use the DDIM sampling strategy for both methods. We do not use any accelerating techniques for a fair comparison. And we note that those techniques used in Stable Diffusion also work in our UniGS.}
\label{tab:param_supp}
\end{table}

\begin{algorithm*}[t!]
\caption{Pseudocode (Python-like) of the Coarse Mask Sampling Method: Curve and Bounding Box}
\label{alg:code}
\definecolor{codeblue}{rgb}{0.25,0.5,0.5}
\definecolor{codekw}{rgb}{0.85, 0.18, 0.50}
\lstset{
  backgroundcolor=\color{white},
  basicstyle=\fontsize{7.5pt}{7.5pt}\ttfamily\selectfont,
  columns=fullflexible,
  breaklines=true,
  captionpos=b,
  commentstyle=\fontsize{7.5pt}{7.5pt}\color{codeblue},
  keywordstyle=\fontsize{7.5pt}{7.5pt}\color{codekw},
}
\begin{lstlisting}[language=python]
prob=random.uniform(0, 1)
## random or bounding box mask

if prob<self.arbitrary_mask_percent:
    mask_img = Image.new('RGB', (W, H), (255, 255, 255))
    bbox_mask=copy.copy(bbox)
    extended_bbox_mask=copy.copy(extended_bbox)
    top_nodes = np.asfortranarray([
                    [bbox_mask[0], (bbox_mask[0]+bbox_mask[2])/2 , bbox_mask[2]],
                    [bbox_mask[1], extended_bbox_mask[1], bbox_mask[1]],
                ])
    down_nodes = np.asfortranarray([
            [bbox_mask[2],(bbox_mask[0]+bbox_mask[2])/2 , bbox_mask[0]],
            [bbox_mask[3], extended_bbox_mask[3], bbox_mask[3]],
        ])
    left_nodes = np.asfortranarray([
            [bbox_mask[0],extended_bbox_mask[0] , bbox_mask[0]],
            [bbox_mask[3], (bbox_mask[1]+bbox_mask[3])/2, bbox_mask[1]],
        ])
    right_nodes = np.asfortranarray([
            [bbox_mask[2],extended_bbox_mask[2] , bbox_mask[2]],
            [bbox_mask[1], (bbox_mask[1]+bbox_mask[3])/2, bbox_mask[3]],
        ])
    top_curve = bezier.Curve(top_nodes,degree=2)
    right_curve = bezier.Curve(right_nodes,degree=2)
    down_curve = bezier.Curve(down_nodes,degree=2)
    left_curve = bezier.Curve(left_nodes,degree=2)
    curve_list=[top_curve,right_curve,down_curve,left_curve]
    pt_list=[]
    random_width=5
    for curve in curve_list:
        x_list=[]
        y_list=[]
        for i in range(1,19):
            if (curve.evaluate(i*0.05)[0][0]) not in x_list and (curve.evaluate(i*0.05)[1][0] not in y_list):
                pt_list.append((curve.evaluate(i*0.05)[0][0]+random.randint(-random_width,random_width),curve.evaluate(i*0.05)[1][0]+random.randint(-random_width,random_width)))
                x_list.append(curve.evaluate(i*0.05)[0][0])
                y_list.append(curve.evaluate(i*0.05)[1][0])
    mask_img_draw=ImageDraw.Draw(mask_img)
    mask_img_draw.polygon(pt_list,fill=(0,0,0))
    mask_tensor=get_tensor(normalize=False, toTensor=True)(mask_img)[0].unsqueeze(0)
    
else:
    mask_img=np.zeros((H,W))
    mask_img[extended_bbox[1]:extended_bbox[3],extended_bbox[0]:extended_bbox[2]]=1
    mask_img=Image.fromarray(mask_img)
    mask_tensor=1-get_tensor(normalize=False, toTensor=True)(mask_img)

\end{lstlisting}
\end{algorithm*}

\paragraph{Referring segmentation}
In the task of referring segmentation, the method of sampling coarse mask remains consistent with the inpainting approach, but the coarse mask region is not excised from the original image. 
Instead, the original image and the coarse mask are concatenated and then jointly input into the model.
Similarly, the model was trained on the COCO and Open-Images dataset for 48 epochs to yield the final model. 
The original image, coarse mask, and text prompt are input during inference.
Post-processing is then performed on the output colormap to obtain the final segmentation mask
 
\paragraph{Entity segmentation}
For entity segmentation, unprocessed original images are input along with an all-one coarse mask, indicating that the entire image area is subject to segmentation. 
During training, we no longer sample entities but directly use all entities from the COCO and EntitySeg datasets, encoding them into a colormap as input for the framework.
In inference, the output colormap undergoes post-processing, but the background is not removed. 
Instead, all clusters are retained as individual entities.

\paragraph{One-to-Many Joint Model} We train our single model for the four tasks within the COCO, Open-Images, and EntitySeg datasets. Unlike training a single model, we add the task embedding for the multi-task training. Furthermore, the sample ratios of four tasks, including multi-class multi-region inpainting, image synthesis, referring segmentation, and entity segmentation, are 0.3, 0.3, 0.2, and 0.2, respectively. We train the single model for multi-tasks in 96 epochs. Finally, this model performs comparably to each model of a single task, indicating the great potential of our unified representation for image generation and segmentation.

\section{More Visualization Results}
\begin{figure}[!tp]
\centering
\includegraphics[height=2.4in,width=0.48\textwidth]{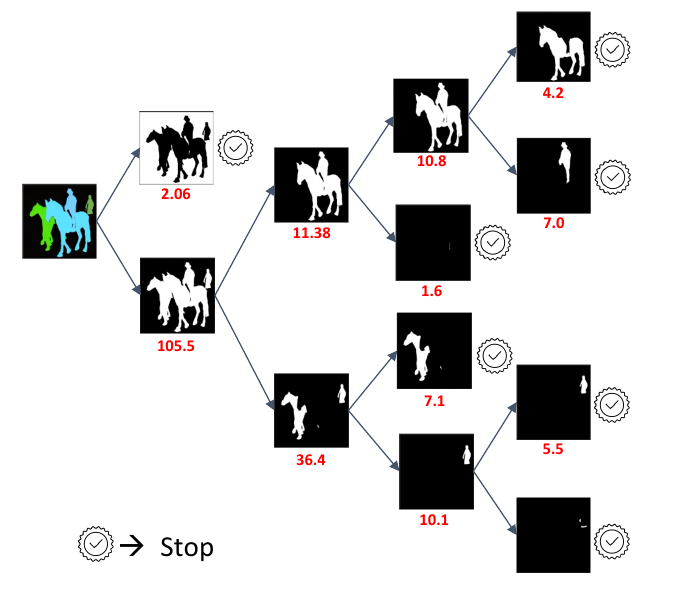}
\vspace{-3mm}
\caption{\textbf{An example illustration of our progressive dichotomy module at each clustering iteration.} The red color indicates the average distance to its cluster center for all the pixels in the cluster.}
\vspace{-4mm}
\label{fig.binary_kmeans}
\end{figure}

\paragraph{Progressive Dichotomy Module.} Figure~\ref{fig.binary_kmeans} shows an example of our progressive dichotomy module to decode the generated colormap into several explicit entity masks. We can see that our decoding process does not require assigning cluster numbers in a depth-first search manner.

\begin{figure*}[!tp]
\centering
\includegraphics[width=0.99\textwidth]{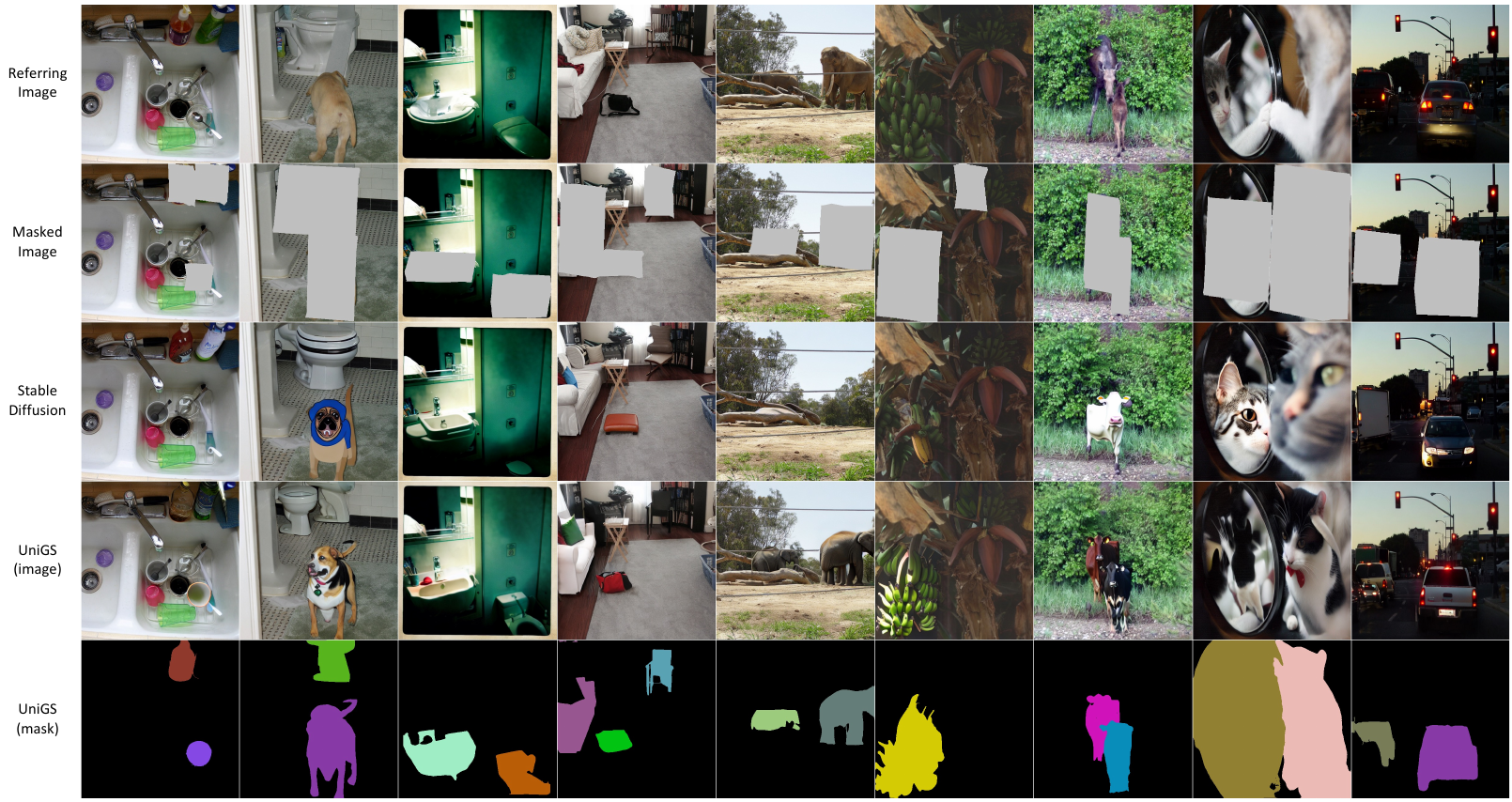}
\vspace{-3mm}
\caption{\textbf{More visualization results of our UniGS in multi-class multi-region inpainting.}}
\vspace{-4mm}
\label{fig.inpainting_supp}
\end{figure*}

\begin{figure*}[!tp]
\centering
\includegraphics[width=0.99\textwidth]{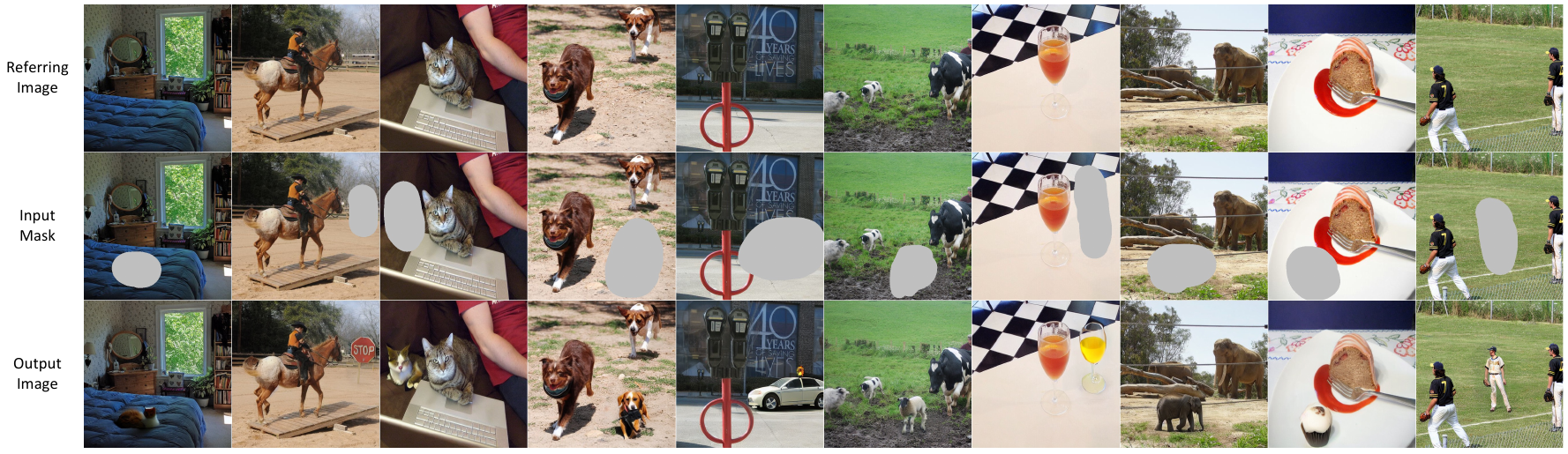}
\vspace{-3mm}
\caption{\textbf{More visualization results of our UniGS in the real world.} We generate the coarse masks used in inpainting by brush strokes from Gradio. We identified some interesting observations, particularly the appearance of shadows from the third to sixth columns.}
\vspace{-4mm}
\label{fig.object_insert_supp}
\end{figure*}

\begin{figure*}[!tp]
\centering
\includegraphics[width=0.99\textwidth]{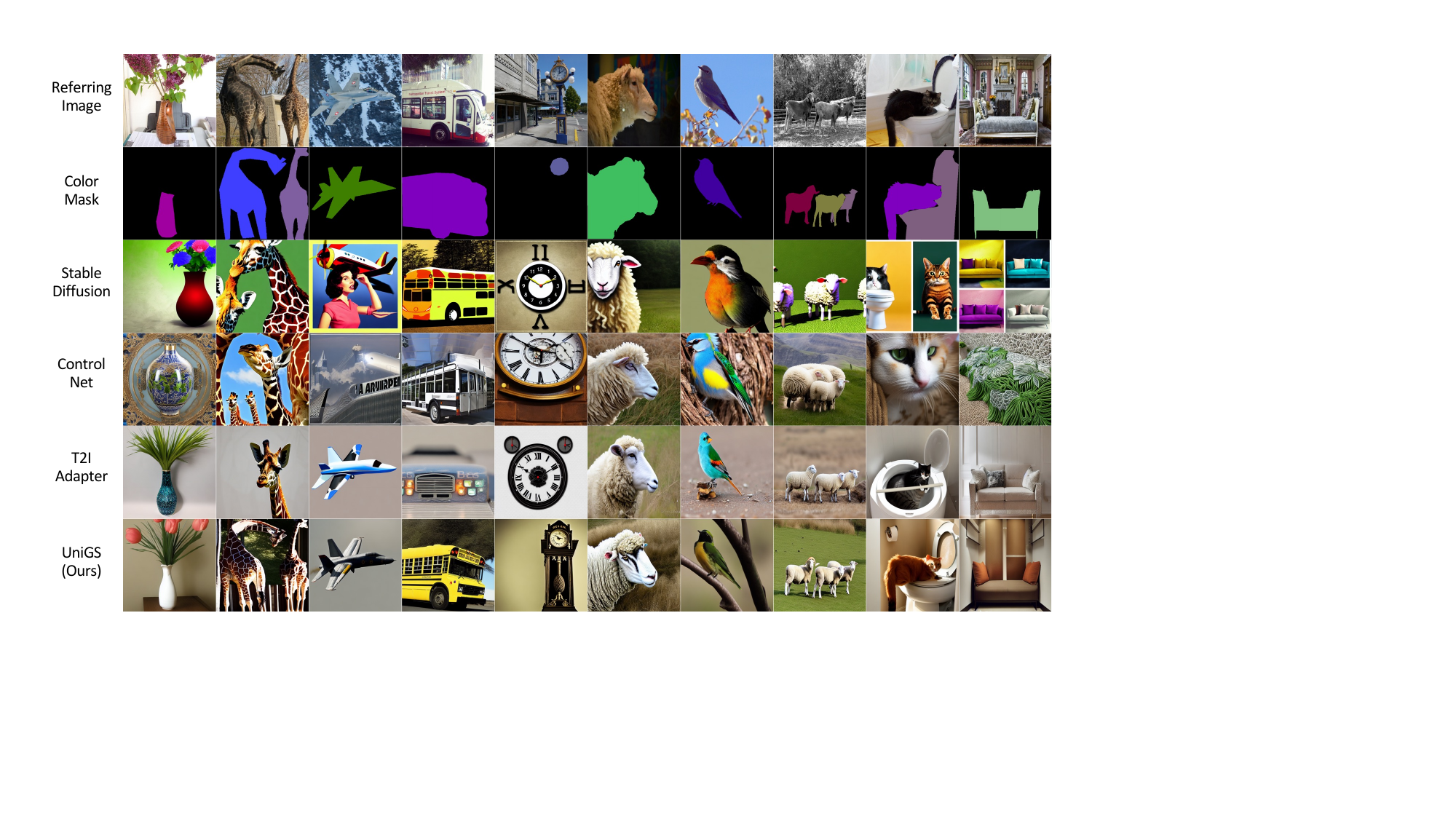}
\vspace{-3mm}
\caption{\textbf{More visualization results of our UniGS in image synthesis}}
\vspace{-4mm}
\label{fig.synthesis_supp}
\end{figure*}

\begin{figure*}[!tp]
\centering
\includegraphics[width=0.99\textwidth]{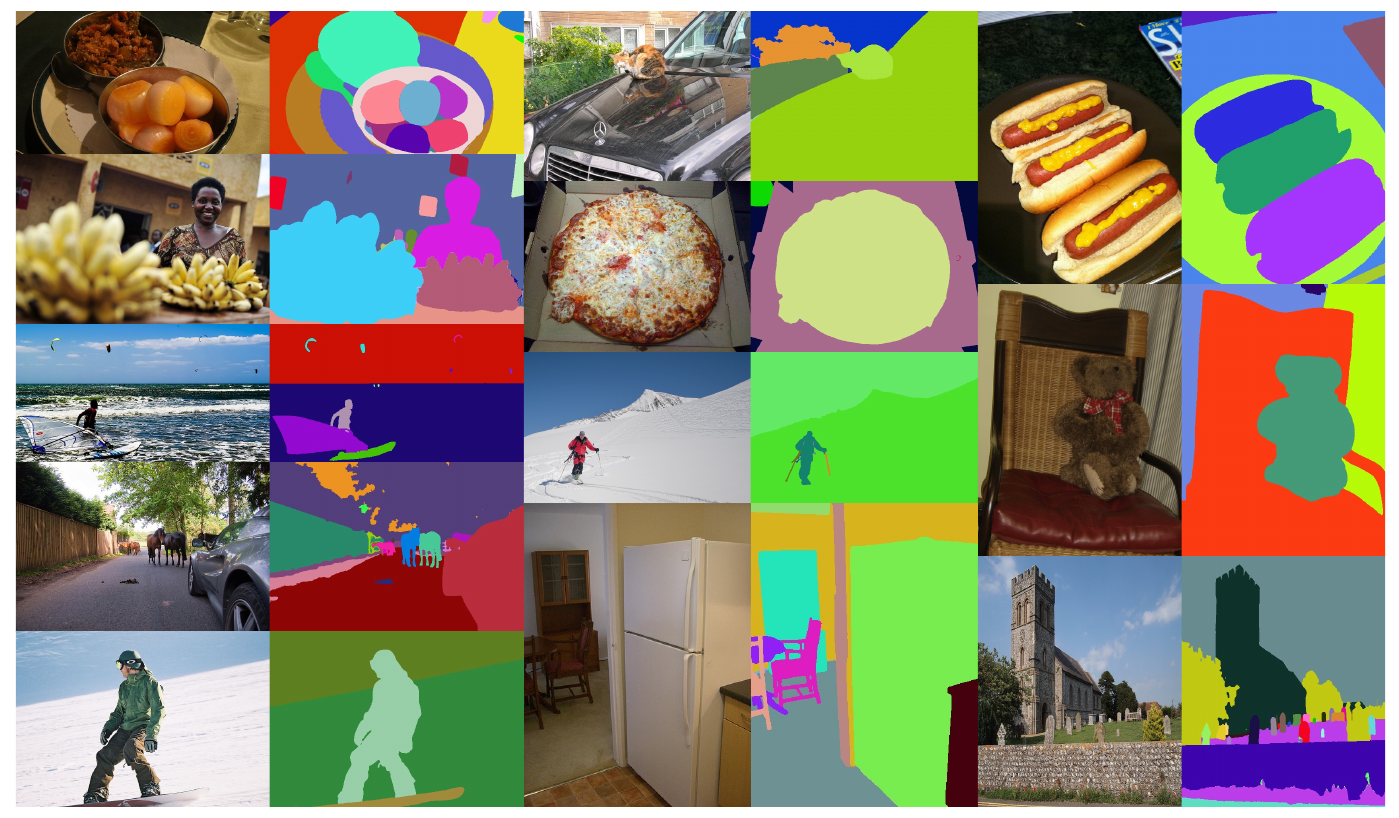}
\vspace{-3mm}
\caption{\textbf{More visualization results of our UniGS in Entity Segmentation.}}
\vspace{-4mm}
\label{fig.entity_supp}
\end{figure*}

\paragraph{Visualization Results}Figure~\ref{fig.inpainting_supp},~\ref{fig.object_insert_supp} and~\ref{fig.synthesis_supp} shows more visualization results of multi-class multi-region inpainting and image synthesis with our UniGS framework. And Figure~\ref{fig.entity_supp} shows the entity segmentation results of our UniGS.


\end{document}